\documentclass[final,x11names,table]{nesy2026} %

\usepackage{longtable}%
\usepackage{multirow}
\usepackage{graphicx}
\usepackage{caption}
\usepackage{csquotes}
\usepackage{amsmath}
\usepackage{amssymb}
\usepackage{mathtools}
\usepackage{multicol}
\usepackage{enumitem}
\newcommand{\g}{\cellcolor{gray!20}}
\usepackage{makecell}
\usepackage{tcolorbox}
\usepackage{enumitem}
\usepackage{listings}
\usepackage{comment}
\usepackage{fvextra}
\usepackage{marginnote}
\usepackage{booktabs}
\usepackage[load-configurations=version-1]{siunitx} %
\usepackage{etoolbox}
\robustify\itshape

\robustify\bfseries
\robustify\g

\newcommand{\entities}{\mathcal{E}}
\newcommand{\prop}{\mathcal{P}}
\newcommand{\attr}{\mathcal{A}}
\newcommand{\clues}{\mathcal{C}}
\newcommand{\option}{\mathcal{S}}
\newcommand{\items}{\mathcal{X}}

\newcommand{\miniexact}{\textsc{MiniExact}}
\newcommand{\methodname}{\textsc{NeuralExactCover}}
\newcommand{\zextreme}{ZebraExtreme}

\DeclareUnicodeCharacter{2208}{$\in$}

\theorembodyfont{\upshape}
\theoremheaderfont{\scshape}
\theorempostheader{:}
\theoremsep{\newline}

\title[From Natural Language to Exact Cover]{From Natural Language to Exact Cover: A Neuro-Symbolic Approach to Zebra Puzzles}%

\author{\Name{Paulius Skaisgiris}\thanks{Equal contribution}\Email{paulius.skaisgiris@tuwien.ac.at} \\
\addr Institute of Logic and Computation, TU Wien, Austria\\[1ex]
\Name{Thomas Pammer}\footnotemark[1] \Email{tpammer@faw.jku.at} \\
\addr Institute for Application-Oriented Knowledge Processing, JKU Linz, Austria\\[1ex]
\Name{Veronika Semmelrock} \Email{veronika.semmelrock@aau.at} \\
\addr Department of Artificial Intelligence and Cybersecurity, University of Klagenfurt, Austria\\[1ex]
\Name{Mykyta Ielanskyi} \Email{ielanskyi@ml.jku.at} \\
\addr Institute for Machine Learning, JKU Linz, Austria\\[1ex]
\Name{Maximilian Heisinger}\thanks{Equal supervision}
\Email{maximilian.heisinger@jku.at} \\
\addr Institute for Symbolic Artificial Intelligence, JKU Linz, Austria \\[1ex]
\Name{Erich Kobler}\footnotemark[2] \Email{erich.kobler@jku.at} \\
\addr Institute for Machine Learning, LIT AI Lab, Department of Virtual Morphology, Clinical Research Institute Medical AI, JKU Linz, Austria
}

\begin{document}

\maketitle

\begin{abstract}
Chain-of-Thought (CoT) generation has substantially improved the performance of Large Language Models (LLMs) on complex reasoning tasks, including code generation, data analysis, and exam-style question answering.
Despite these advances, purely neural LLMs continue to struggle with elementary logical reasoning problems and lack the determinism, soundness, and reliability characteristic of symbolic reasoning systems.
Conversely, classical symbolic methods such as SAT solving and exact cover guarantee correctness and completeness, but require problems to be expressed in highly specialized formal encodings, limiting their applicability to natural language inputs.
In this work, we present a synergistic neuro-symbolic framework that bridges this gap by combining neural semantic parsing with deterministic exact cover reasoning.
Given a Zebra-style logic puzzle in natural language, an LLM translates entities and constraints into structured tool calls that construct an exact cover instance.
A symbolic solver then computes the solution and provides feedback to iteratively refine invalid, under- or overconstrained encodings, exploiting the uniqueness property of Zebra puzzles.
We conduct a comprehensive empirical evaluation across multiple parameter scales, post-training paradigms, and LLM families.
The results demonstrate that our hybrid approach consistently outperforms strong plain neural baselines, including CoT prompting, as well as recent neuro-symbolic methods.\footnote{Code is available at \url{https://codeberg.org/bilai-ss25-booleguru-llm/zebrallm}.}
\end{abstract}

\section{Introduction}
\label{sec:intro}

Constraint Satisfaction Problems (CSPs) lie at the core of many real-world decision-making tasks, including scheduling, planning, resource allocation, and configuration, making them a fundamental abstraction in AI research~\citep{Russel:1995,Mackworth:1977,Nadel:1989,donald2016art,donald2020art,donald2025art}.
While symbolic methods such as SAT and constraint solvers can guarantee correctness and completeness, their practical deployment critically depends on constructing effective formal encodings -- a process that remains difficult even for domain experts~\citep{DBLP:conf/ijcar/HeisingerHS24}.
Translating informal problem descriptions into a structured encoding requires substantial modeling expertise as small representational choices can dramatically impact solver performance and tractability.
As a result, research often studies simplified yet representative proxy domains such as logic puzzles (e.g. Sudoku or Zebra Puzzles), which capture essential combinatorial and relational reasoning challenges while providing clean semantics, controllable complexity, and objectively verifiable solutions \citep{Dechter:1987,Prosser:1993,lin2025zebralogic,whiteLiveBenchChallengingContaminationLimited2025}.

This reliance on structured benchmark domains aligns naturally with recent advances in Large Language Models (LLMs), which are increasingly applied to tasks requiring the construction of structured symbolic solutions from unstructured inputs.
Such domains include mathematical reasoning~\citep{Hendrycks:2021a}, programming~\citep{Hendrycks:2021}, planning~\citep{Zhang:2026}, and logic puzzle solving~\citep{lin2025zebralogic,Chollet:2025}.
A shared characteristic of these problems is that, although discovering solutions is challenging, their correctness can be verified algorithmically.
Reinforcement learning with verifiable rewards post-training has therefore emerged as a standard approach for optimizing models directly against symbolic verifiers~\citep{DeepSeek-AI:2025}.
The resulting thinking LLMs produce Chain-of-Thought (CoT) reasoning traces enabling inference-time scaling.

\begin{figure}[t]
    \centering
    \includegraphics[width=0.9\linewidth]{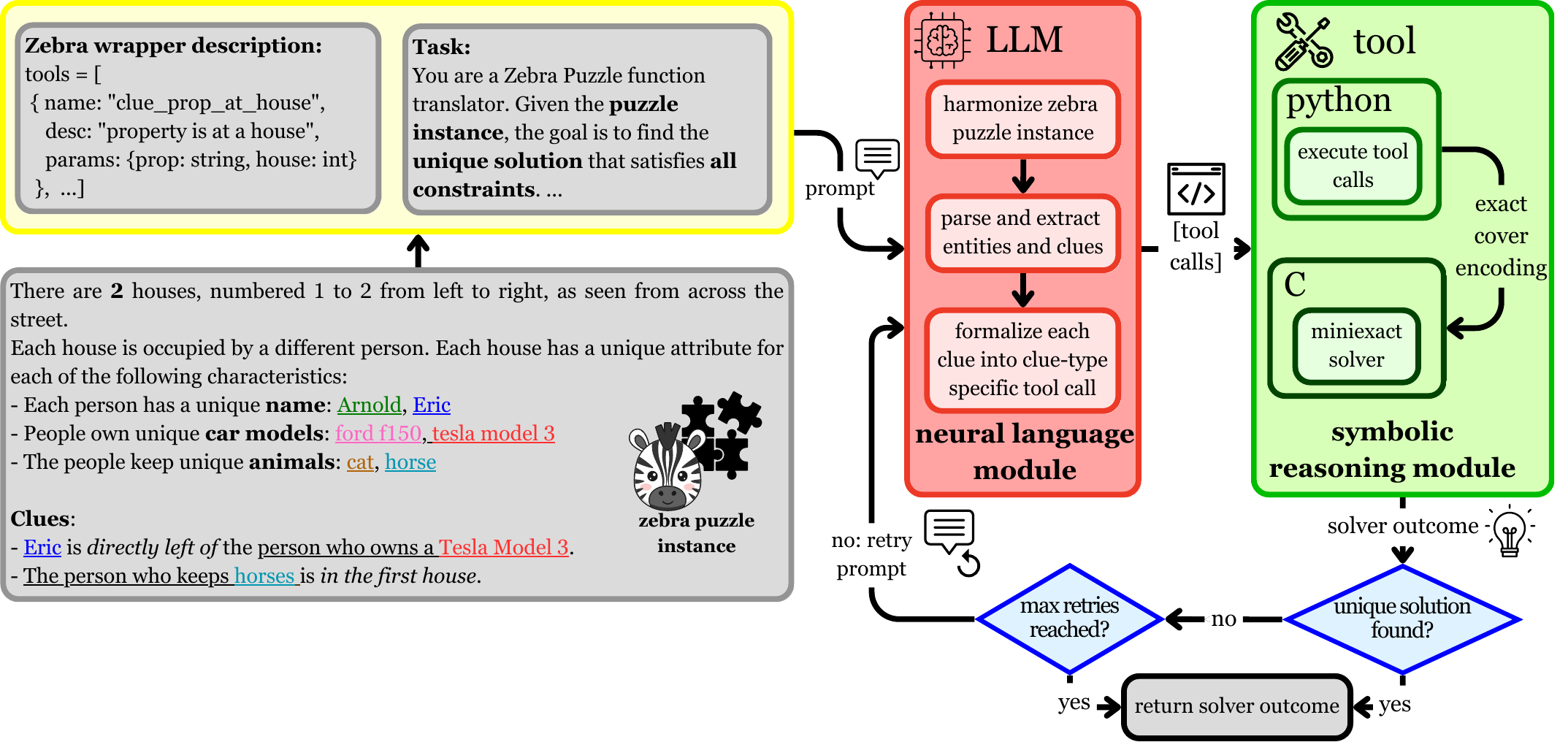}
    \caption{
    Visualization of \methodname{}. A puzzle instance is combined with our prompt template and processed by an LLM that harmonizes attributes, extracts entities and clues, and translates clues into an exact cover encoding.
    The symbolic \miniexact{} solver then solves the encoding and provides iterative feedback on failures to refine the generated tool calls. This loop continues until a unique solution is found or the retry limit is reached, after which the solution is converted into plain text.
    }
    \label{fig:main}
\end{figure}

Zebra Puzzles are among recent benchmarks to test the reasoning capabilities of LLMs \citep{lin2025zebralogic, whiteLiveBenchChallengingContaminationLimited2025}.
They consist of lists of entities, properties, and attributes as well as clues; the goal is to infer the relations of attributes and entities from the given information.
\citet{lin2025zebralogic} find that scaling the model size does not help beyond a certain level of complexity which is defined by the puzzles' solution space. 
This manifests in generalist LLMs faring extremely well on simple Zebra Puzzles while struggling with more complex ones.
Longer reasoning traces appear to be more helpful with these problems than larger models.
However, these longer traces present additional challenges such as high frequency noise \citep{Radler:2026} and increased chance of hallucination \citep{Mei:2025}.

To address the reasoning limitations of purely neural models, recent neuro-symbolic approaches increasingly combine LLM-based semantic parsing with symbolic solvers to enforce logical consistency \citep{Dinu:2024}.
Prior work explores both agentic formulations that iteratively generate solver-compatible programs and solver-centric pipelines that translate natural language inputs into formal constraint specifications solved using SAT, SMT, CSP, or ASP techniques~\citep{szeider2025cp,pan2023logic,ye2023satlm,pmlr-v284-eiter25a}.
Notably, Logic.py~\citep{kesseli2025logic} translates natural language into a domain-specific representation solved by an exact constraint solver, incorporating iterative error feedback to the LLM.
This coupling of semantic parsing and symbolic solving consistently outperforms LLMs on Zebra Puzzle-style benchmarks.

In this work, we present an integrated neuro-symbolic framework called \methodname{} that combines LLM-based semantic parsing with deterministic exact cover solving for reliable logical reasoning over plain text Zebra-style puzzles.
Our main contributions are as follows:
\begin{itemize}[leftmargin=1em,noitemsep,partopsep=0pt,topsep=0pt,parsep=0pt]
    \item We introduce and adapt an exact cover encoding for efficient symbolic solving of Zebra-style logic puzzles.
    \item We introduce a neuro-symbolic framework that leverages LLMs as a translation layer for an exact cover solver and re-applies symbolic results as feedback to the LLM.
    \item We introduce the \zextreme{} dataset, extending ZebraLogic~\citep{lin2025zebralogic} with even more challenging problems of size up to $8\times10$ entities and properties.
\end{itemize}

\section{Background and Preliminaries}\label{sec:background}

\subsection{Exact Cover}

Given a collection $S$ of subsets of a set $\items$, the \emph{exact cover problem} is about finding a subcollection $\option^*\subseteq{\option}$ such that each element in $\items$ is contained in \emph{exactly one} subset of $\option^\ast$.
This concept becomes clearer when we rename elements in $\items$ to \emph{items} and $\option$ to \emph{options}.
We want to find the subcollection $\option^*$ of options to cover all items exactly once.
One example comes from~\cite{donald2020art}: Let~$\items=\{a,b,c,d,e,f,g\}$ be items and the collection of options $\option=\{{\bigl[c,e\bigr],\bigl[a,d,g\bigr],\bigl[b,c,f\bigr],\bigl[a,d,f\bigr],\bigl[b,g\bigr],\bigl[d,e,g\bigr]}\}$.
When selecting the first, fourth, and fifth options, we cover each item \emph{exactly} once, i.e.~$\option^\ast=\{\bigl[c,e\bigr],\bigl[a,d,f\bigr],\bigl[b,g\bigr]\}$.
This is very useful, as such exact-1 properties have a large variety of encodings with strong performance impacts for solvers, as discussed in the context of SAT-solving in~\cite{10.1145/3426020.3426170}.

\paragraph{Colored Exact Cover.} Exact cover can be extended by splitting the options into \emph{primary items}
$\items_P$ and \emph{secondary items} $\items_S$ such that $\items=\items_P \cup \items_S$.
Let $\entities$ denote a set of colors that can be used to \emph{color} (change) secondary items.
Now, each primary item must still be covered \emph{exactly once}, while secondary items must be assigned \emph{at most one} color.
Note that colors can be shared between options, i.e.~multiple options with secondary items of matching colors may be selected.

\paragraph{Notation.} We write options as sets of comma-separated items enclosed in brackets, e.g., $\bigl[a,b\bigr]$.
We use colons to separate secondary items and colors.
Thus, $a:c$ means the secondary item $a\in\items_S$ has the color $c\in\entities$.
Items may be combined from other terms using angle brackets, e.g. $\langle{}a,b\rangle{}$ is one item.
For instance, $\bigl[\langle{}a,b\rangle{}:{}1, c\bigr]$ denotes an option of one item $\langle{}a,b\rangle{}$ with the color $1$ (meaning $\langle{}a,b\rangle{}$ must be a secondary item), and $c$ as another item.
A set of options is enclosed in curly brackets.

\paragraph{\miniexact{}.} It is an efficient implementation\footnote{\url{https://github.com/miniexact/miniexact}} of some of the Dancing Links algorithms of~\citet{donald2020art}.
Its performance mostly mirrors Knuth's original C implementation, while also offering a user-friendly Python API and a more sophisticated parser.
The algorithm can be selected at run-time, making it a useful tool for experiments with exact cover problems.

\subsection{Zebra Puzzles}
Zebra Puzzles are classic constraint satisfaction problems (CSPs) where a solver must find a one-to-one mapping across entity categories using logical clues (see Fig.~\ref{fig:main}).

\paragraph{Formulation.} Each puzzle consists of initial statements defining a finite set of entities
$\entities = \{e_1, \ldots, e_N\}$ (e.g., numbered houses) with
$N>1$; several characteristics or properties~$\prop=\{p_1,\ldots,p_M\}$ (e.g., name, car, animal) with $M\geq1$, each containing $N$ mutually exclusive attributes; and a set of clues $\clues = \{c_1, \ldots, c_K\}$ with $K\geq1$ that relate entities,
properties, and attributes.
All attributes of a property define its domain $\attr_p=\{a_{p1},\ldots,a_{pN}\}$ (e.g., cat, horse) with $\vert\attr_p \vert=N$.

\paragraph{Constraints.} To ensure uniqueness of each puzzle's solution, \emph{uniqueness} and \emph{clue-based} constraints have to be fulfilled.
The \emph{uniqueness constraint} ensures that every attribute $a\in\attr_p$ is assigned exactly to one entity for all $p\in\prop$.
\emph{Clue-based constraints} formalize the different verbal clues into logic statements that must be satisfied by a valid solution.
For Zebra Puzzles, there are 8 different clue-based constraints that relate attributes and/or entities by specifying presence/absence or positional relations as summarized in Appx.~\ref{appx:clue-encodings}.

\paragraph{Task.} The goal is to find an assignment $x_{p,e}\in\attr_p$ selecting an attribute for each property $p\in\prop$ and entity $e\in\entities$ such that all constraints are fulfilled.
Note that these assignments can be summarized into a tabular grid where the columns are defined by the entities, the rows by the properties, and the assigned attributes are listed in the corresponding cells.

\paragraph{Datasets.} Zebra Puzzles are used as proxies to test the reasoning capabilities of modern LLMs \citep{lin2025zebralogic, whiteLiveBenchChallengingContaminationLimited2025}.
To this end, \citet{lin2025zebralogic} introduced the ZebraLogic dataset containing 1,000 logic grid puzzles of various sizes ($N\times M$) ranging from $2\times2$ to $6\times6$.
We extend this dataset to include 400 bigger puzzle instances ($7\times5$, $\ldots$, $8\times10$) and call this extended dataset \zextreme\footnote{Data is available at \url{https://huggingface.co/datasets/pskaisg/ZebraExtreme}.}.
Following \citet{lin2025zebralogic}, we partition the puzzles according to their solution space size ($\vert\mathcal{Q}\vert=(N!)^M$) into bins:
\begin{align*}    \mathrm{S}&\colon\;10^{0}\leq\vert\mathcal{Q}\vert<10^{3}, 
&\mathrm{M}&\colon\;10^{3}\leq\vert\mathcal{Q}\vert<10^{6},
&\mathrm{L}&\colon\;10^{6}\leq\vert\mathcal{Q}\vert<10^{10},\\
\mathrm{XL}&\colon\;10^{10}\leq\vert\mathcal{Q}\vert<10^{18}, 
&\mathrm{XXL}&\colon\;10^{18}\leq\vert\mathcal{Q}\vert<10^{30},
&\mathrm{XXXL}&\colon\;10^{30}\leq\vert\mathcal{Q}\vert<10^{50}.
\end{align*}
Note that the bins S to XL are included in the original ZebraLogic dataset, while the \zextreme{} only includes puzzles in XXL and XXXL.
Appx.~\ref{appx:zebra-extreme} describes how we created this computationally challenging dataset.

\section{Solving Zebra Puzzles with Exact Cover}
In the following, we present \methodname{}, a neural-symbolic approach for solving Zebra logic puzzles.
As in Figure~\ref{fig:main}, we leverage the in-context learning capabilities of LLMs to extract semantic relations between entities, properties, and attributes from free-text puzzle descriptions and translate them into an exact cover encoding.
The symbolic module then computes solutions and provides feedback on the resulting encoding, allowing the LLM to iteratively refine its tool calls until a unique solution is obtained.
This feedback loop integrates neural language understanding with symbolic reasoning.

\subsection{Language Module for Puzzle Parsing}

The main goal of the language module is the conversion of the problem from natural language to a valid exact cover encoding.
At the core of this module is a language model.
To this end, we consider several recent open weight LLMs that are state-of-the-art for their parameter size, such as Qwen3 \citep{Yang:2025}, gpt-oss \citep{OpenAI:2025}, Magistral \citep{Magistral:2025}, and Devstral \citep{Devstral:2025}.
Note that any language model that has been fine-tuned to perform tool calling could be used.
The model is prompted to complete the following sequence of actions: \emph{harmonization}, \emph{clue extraction} and \emph{formalization}.
The full prompt is listed in Appx.~\ref{appx:prompt-used}.
It includes manually constructed few-shot examples to improve the adherence of the model to the specific instructions given. 

\paragraph{Harmonization.} First, each puzzle is paraphrased such that every attribute occurrence follows the format  $<$property-name$>$@$<$attribute-name$>$.
This reduces ambiguity arising from attributes that occur under multiple properties.
For example, if ``cat'' is used by two different properties, it may be extracted incorrectly, resulting in an erroneous encoding.

\paragraph{Clue extraction.} First, the LLM parses the puzzle's setup (i.e.~$M$, $N$, $\{\attr_p\}_{p\in\prop}$, and $K$) and proceeds to extract the clues. 
The model is instructed to use our exact cover encoding's tool-calls presented in the next section to express the extracted clues.
Consequently, the output of the LLM is a list of tool calls in JSON format, specifying the called function and its arguments.
The first call must be to the initialization tool, that specifies the dimensions of the problem, i.e. number of entities, properties, property domains, and clues.
The subsequent tool calls encode each clue.
The complete reference of callable tools is provided in Appx.~\ref{appx:miniexact-impl-details}.
Note that we prompt each model to provide a one-sentence description for each tool call to simplify the optional inspection of traces (Appx.~\ref{appx:traces}).

\subsection{Reasoning Module using Exact Cover}

We use exact cover as a logic backbone since it natively expresses
exact-one constraints instead of requiring complex encodings as in
SAT-solving and solvers naturally support enumerating
  solutions without employing blocking
  clauses~\citep{10.1145/3426020.3426170}.

\paragraph{Items.} We assign a primary item to all clues~$\clues$, ensuring that each must be used once.
We expand entities and properties over their full domain and add a primary item for each pair, to model that each property has exactly one entity.
Moreover, we use entities to color secondary items, defined by all attributes of all properties.
Thus, the set of items is
\[
\items=\items_P \cup \items_S,\quad
\items_P = \clues \cup \left(\bigcup_{e\in\entities} \bigcup_{p\in\prop} \Bigl\{\big[\langle{}p,e\rangle{}\big]\Bigr\}\right), \quad 
\items_S = \bigcup_{p\in\prop} \bigcup_{a\in\attr_p} \Bigl\{\big[a\big]\Bigr\}.
\]

\paragraph{Options.} To model the uniqueness constraint -- every attribute of each property can only be assigned to one entity, we set
\[
\option_U = \bigcup_{e \in \entities}\bigcup_{p\in\prop} \bigcup_{a\in\attr_p} \Bigl\{\big[\langle{}p,e\rangle{},a: e\big]\Bigr\}.
\]
The clue-based constraints are transformed into corresponding exact cover encodings based on the 8 different clue types.
Then, the set of options reads as
\[
\option = \option_U \cup \option_\clues, \quad \option_\clues = \bigcup_{c\in\clues} \option_c,
\]
where $S_c$ implements the encoding of clue~$c$ according to its clue
type listed in Appx.~\ref{appx:clue-encodings}. Because each clue has a
primary item, exactly one of the clue's options must be in the
solution. This forces the secondary items representing the affected
attributes to be colored according to the clue, in turn affecting
other clues and the solution as a whole.

\paragraph{Implementation Details.}
We implement the proposed exact cover encoding using the \miniexact{} Python API.
The encoding is exposed through a set of documented functions corresponding to the clue types in Appx.~\ref{appx:clue-encodings}, enabling LLMs to construct exact cover instances via tool calls.
To ensure robust evaluation, extracted property and attribute names are normalized to canonical puzzle names using string-similarity matching.
Additional implementation details and API descriptions are provided in Appx~\ref{appx:miniexact-impl-details}.

\subsection{Feedback Mechanism}
\label{sec:feedback}

For neuro-symbolic integration, \methodname{} uses feedback from the symbolic reasoning component to iteratively refine the LLM-generated tool calls.
First, the generated tool calls are checked for completeness by verifying that tool calls were produced and that their number matches the number of puzzle clues.
If tool calls are missing or excessive, the LLM is prompted to generate one tool call per clue, add missing clue encodings, or re-evaluate redundant or conflicting ones.

If the number of tool calls and parsed clues matches, the corresponding encoding is evaluated by the \miniexact{} solver.
For feedback, the following three outcomes are considered:
\begin{description}[leftmargin=1em,topsep=0ex,parsep=0ex,itemsep=0ex]
\item[Overconstrained] 
If \emph{no solution} exists, the generated constraints are overconstraining, indicating an error in clue extraction.
The LLM receives symbolic feedback and is instructed to revise at least one previous tool call to make the encoding less restrictive.
\item[Underconstrained]
If \emph{multiple solutions} are found, the encoding is underconstrained, and the LLM is prompted to revise its clue interpretations by modifying at least one tool call.
\item[Unique solution]
If a \emph{unique solution} is found, the encoding is considered sufficiently complete and consistent, and the solution is returned.
\end{description}
The corresponding prompts and additional details are listed in Appx.~\ref{appx:feedback_prompt}.

\section{Experiments}
We evaluate the reasoning performance of plain LLMs (without tools), Logic-LM~\citep{pan2023logic}, Logic.py~\citep{kesseli2025logic}, and our \methodname{} (NEC) in a LLM model agnostic way on Zebra Puzzles.
To apply Logic-LM to these logic puzzles, we extended its CSP pipeline by adding a Zebra puzzle specific prompt and adapting the solution parsing, see Appx~\ref{appx:logic-lm}.
Note that Logic.py implements a feedback mechanism to correct for syntax errors and UNSAT encodings.
For numerical evaluation, we consider the recent ZebraLogic benchmark~\citep{lin2025zebralogic} and our introduced \zextreme{}, a more challenging variant characterized by a larger number of cells and an expanded set of possible values per cell.
Refer to Appx~\ref{appx:zebra-extreme} for details on this dataset.

To assess the reasoning and scaling capabilities of the different methods, we evaluate puzzle and cell accuracy.
While the former counts the fraction of correctly solved puzzles, the latter counts the fraction of correctly assigned attributes to entities (i.e.~cells in the solution table).
In addition, we evaluate the runtime of the LLMs and the symbolic solvers.
We stratify the runtime and puzzle accuracy by the puzzle size into bins S to XXXL to analyze how models perform on increasingly complex puzzles.

For all methods, each LLM was run with \texttt{temperature} set to \num{0} and \texttt{top\_p} set to \num{1} following \cite{lin2025zebralogic}. 
We also adopt an output token budget of 4096 per LLM call, except for \zextreme{}, where the budget is increased to 30720 to accommodate reasoning over larger puzzles.
Notably, Logic.py~\cite{lin2025zebralogic} employs a multi-step LLM pipeline checking for syntax and constraint-encoding errors at each stage and reruns LLMs when such errors occur, thereby enabling multiple self-correction attempts. Consequently, the total token usage may exceed the output token budget of 4096.
In contrast, our approach relies on a single LLM call.
Further details on models and serving are in Appx~\ref{appx:models}.
Since all four approaches under consideration are LLM-agnostic, we evaluate LLMs across a diverse range of parameter scales, training paradigms, and families.

\begin{table}[bt!]
\sisetup{
  detect-weight=true,
  detect-family=true,
  table-number-alignment=center,
  table-format=3.2,
  parse-numbers=true,
  round-mode=places,
  round-precision=2,
  round-pad=false,
}
\caption{
Effect of training mode of Qwen3 LLMs. The best result per model-size group is shown in \textit{italics}. The best result across all models is shown in \textbf{bold italics}.
}
\scriptsize
\setlength{\tabcolsep}{3pt}
\resizebox{\textwidth}{!}{%
\begin{tabular}{clSSSSSSSSSSSSSSS}
\toprule
Size & Metric & \multicolumn{5}{c}{Base} & \multicolumn{5}{c}{Instruct} & \multicolumn{5}{c}{Thinking} \\
\cmidrule(r{0.3em}){3-7} \cmidrule(r{0.3em}){8-12} \cmidrule(r{0.3em}){13-17}
 &  & \multicolumn{1}{c}{P} & \multicolumn{1}{c}{LL} & \multicolumn{1}{c}{\cellcolor{gray!20}NEC} & \multicolumn{1}{c}{LP$^\circlearrowright$} & \multicolumn{1}{c}{\cellcolor{gray!20}NEC$^\circlearrowright$} & \multicolumn{1}{c}{P} & \multicolumn{1}{c}{LL} & \multicolumn{1}{c}{\cellcolor{gray!20}NEC} & \multicolumn{1}{c}{LP$^\circlearrowright$} & \multicolumn{1}{c}{\cellcolor{gray!20}NEC$^\circlearrowright$} & \multicolumn{1}{c}{P} & \multicolumn{1}{c}{LL$^\dagger$} & \multicolumn{1}{c}{\cellcolor{gray!20}NEC} & \multicolumn{1}{c}{LP$^\circlearrowright$} & \multicolumn{1}{c}{\cellcolor{gray!20}NEC$^\circlearrowright$} \\
\midrule
\multirow{6}{*}{4B} & All & 52.7 & 25.6 & \multicolumn{1}{c}{\g 82.6} & 68 & \multicolumn{1}{c}{\g \itshape 84.9} & 66.1 & 43.5 & \multicolumn{1}{c}{\g 53.8} & 80.2 & \multicolumn{1}{c}{\g 65.7} & 27.5 & \textendash & \multicolumn{1}{c}{\g 3.3} & 45.1 & \multicolumn{1}{c}{\g 10.3} \\
 & S & \multicolumn{1}{c}{\itshape 98.44} & 25.94 & \multicolumn{1}{c}{\g 95} & 82.5 & \multicolumn{1}{c}{\g 96.56} & 97.81 & 70.31 & \multicolumn{1}{c}{\g 71.88} & 93.12 & \multicolumn{1}{c}{\g 82.19} & 79.06 & \textendash & \multicolumn{1}{c}{\g 9.38} & 55.31 & \multicolumn{1}{c}{\g 29.38} \\
 & M & 71.07 & 26.43 & \multicolumn{1}{c}{\g 93.93} & 72.5 & \multicolumn{1}{c}{\g \itshape 96.79} & 85.71 & 50.71 & \multicolumn{1}{c}{\g 57.14} & 85.36 & \multicolumn{1}{c}{\g 73.57} & 7.86 & \textendash & \multicolumn{1}{c}{\g 1.07} & 44.29 & \multicolumn{1}{c}{\g 3.21} \\
 & L & 6.5 & 24 & \multicolumn{1}{c}{\g 82} & 65.5 & \multicolumn{1}{c}{\g \itshape 83} & 48 & 27 & \multicolumn{1}{c}{\g 39.5} & 73.5 & \multicolumn{1}{c}{\g 52.5} & 0 & \textendash & \multicolumn{1}{c}{\g 0} & 36 & \multicolumn{1}{c}{\g 0} \\
 & XL & 0 & 25.5 & \multicolumn{1}{c}{\g 47.5} & 41 & \multicolumn{1}{c}{\g 51.5} & 6 & 7 & \multicolumn{1}{c}{\g 34.5} & \multicolumn{1}{c}{\itshape 59} & \multicolumn{1}{c}{\g 41.5} & 0 & \textendash & \multicolumn{1}{c}{\g 0} & 39 & \multicolumn{1}{c}{\g 0} \\
 & Cell & 33.65 & 28.47 & \multicolumn{1}{c}{\g 76.19} & 73.48 & \multicolumn{1}{c}{\g 77.61} & 52.41 & 50.21 & \multicolumn{1}{c}{\g 55.5} & \multicolumn{1}{c}{\itshape 84.98} & \multicolumn{1}{c}{\g 63.81} & 13.54 & \textendash & \multicolumn{1}{c}{\g 1.47} & 42.91 & \multicolumn{1}{c}{\g 4.8} \\
\midrule
\multirow{6}{*}{30B} & All & 54.2 & 31.1 & \multicolumn{1}{c}{\g 93.6} & 93.3 & \multicolumn{1}{c}{\g \bfseries\itshape 95.6} & 70.9 & 83 & \multicolumn{1}{c}{\g 92.7} & 87.5 & \multicolumn{1}{c}{\g 95} & 35.5 & \textendash & \multicolumn{1}{c}{\g 30.2} & 86.5 & \multicolumn{1}{c}{\g 39.4} \\
 & S & 99.06 & 18.12 & \multicolumn{1}{c}{\g 97.81} & 98.75 & \multicolumn{1}{c}{\g 97.81} & \multicolumn{1}{c}{\bfseries\itshape 99.38} & 95.31 & \multicolumn{1}{c}{\g 95.94} & 95.94 & \multicolumn{1}{c}{\g 98.44} & 92.19 & \textendash & \multicolumn{1}{c}{\g 69.69} & 87.81 & \multicolumn{1}{c}{\g 87.5} \\
 & M & 75 & 29.29 & \multicolumn{1}{c}{\g 97.5} & 94.64 & \multicolumn{1}{c}{\g \bfseries\itshape 98.57} & 95.71 & 87.14 & \multicolumn{1}{c}{\g 95} & 88.57 & \multicolumn{1}{c}{\g 97.5} & 21.43 & \textendash & \multicolumn{1}{c}{\g 23.93} & 91.07 & \multicolumn{1}{c}{\g 36.79} \\
 & L & 7.5 & 42 & \multicolumn{1}{c}{\g 93.5} & 93 & \multicolumn{1}{c}{\g \bfseries\itshape 95.5} & 55 & 74.5 & \multicolumn{1}{c}{\g 91.5} & 85 & \multicolumn{1}{c}{\g 92.5} & 0 & \textendash & \multicolumn{1}{c}{\g 6} & 85.5 & \multicolumn{1}{c}{\g 5.5} \\
 & XL & 0 & 43.5 & \multicolumn{1}{c}{\g 81.5} & 83 & \multicolumn{1}{c}{\g 88} & 6.5 & 66 & \multicolumn{1}{c}{\g 85.5} & 75 & \multicolumn{1}{c}{\g \bfseries\itshape 88.5} & 0 & \textendash & \multicolumn{1}{c}{\g 0} & 79 & \multicolumn{1}{c}{\g 0} \\
 & Cell & 34.79 & 37.64 & \multicolumn{1}{c}{\g 92.64} & 93.39 & \multicolumn{1}{c}{\g \bfseries\itshape 95.81} & 54.28 & 85.67 & \multicolumn{1}{c}{\g 93.77} & 85.44 & \multicolumn{1}{c}{\g 95.54} & 19.12 & \textendash & \multicolumn{1}{c}{\g 16.76} & 87.67 & \multicolumn{1}{c}{\g 22.82} \\
\bottomrule
\end{tabular}
}
{\footnotesize $\circlearrowright$ denotes approaches using a feedback loop.
P: Plain; LL: Logic-LM; NEC: NEC~{\tiny (ours)}; LP: Logic.py.\\
{\footnotesize$^\dagger$ A parsing error prevented this configuration from returning any output.}}
\label{tab:RQ2}
\end{table}

\paragraph{LLM Post-Training Paradigms.} We examined the effect of instruction following and thinking post-training paradigms on solving Zebra puzzles.
To this end, we evaluated the Qwen3 model family in the hybrid (Base), instruct, and thinking configuration both at 4B and 30B parameter scales.
Table~\ref{tab:RQ2} lists the corresponding puzzle and cell accuracies.
For small puzzles, the Plain approach consistently achieved high performance -- most results exceeding $90\%$, but performance dropped sharply on larger puzzles. 
In contrast, Logic.py attained lower accuracy on small puzzles, yet its performance degraded more gradually as puzzle size increased, while Logic-LM only achieved comparable accuracy for 30B instruct.
Our approach outperformed all neurosymbolic baselines in all settings except 4B instruct.

Interestingly, for Plain, Logic-LM and Logic.py the instruct post-training strategy outperforms the thinking versions. 
We hypothesize that, for these approaches, extended reasoning may lead the model to focus on irrelevant details and become confused, whereas instruct variants remain more focused on producing a solution.

For \methodname{}, the hybrid variant outperforms both the instruct and thinking paradigms. We attribute this to its ability to combine concise, goal-directed tool usage with explicit step-by-step reasoning. Since the model is not required to solve the puzzle directly, but instead to encode it as an exact cover problem through appropriate tool calls, the hybrid setup appears particularly effective, as it allows the model to reason over individual clues while still producing a structured and executable formulation of the full problem.

\paragraph{Performance across LLM families.} 
We compare the performances of the five considered approaches with various LLM families in the $20$B-$30$B range (details on models used can be found in Appx.~\ref{appx:models}).
The results are presented in Tab.~\ref{tab:RQ3_combined}.
Both versions of \methodname{} perform well across most tested LLMs, where the feedback-loop approach consistently outperforms the approach using a single run, often achieving highest puzzle and cell accuracies on ZebraLogic. In particular, it attains the best overall performance with Qwen3-30B-Instruct and Devstral Small-2, outperforming the compared approaches.

At the time of writing, the official ZebraLogic SOTA is 91.7\% puzzle and 95.7\% cell accuracy by o3-mini-2025-01-31-high\footnote{Leaderboard accessed 2026-06-13 at \url{https://huggingface.co/spaces/allenai/ZebraLogic}.} (Plain approach).
Our \methodname{} using the feedback loop with mistralai/Devstral-Small-2-24B-Instruct-2512 achieves 97.9\% puzzle and 98.32\% cell accuracy, establishing a new SOTA.
We also provide the first \zextreme{} baseline, reaching 84.75\% puzzle and 82.59\% cell accuracy using gpt-oss-20b.

\begin{figure}
    \centering
    \includegraphics[width=.95\linewidth]{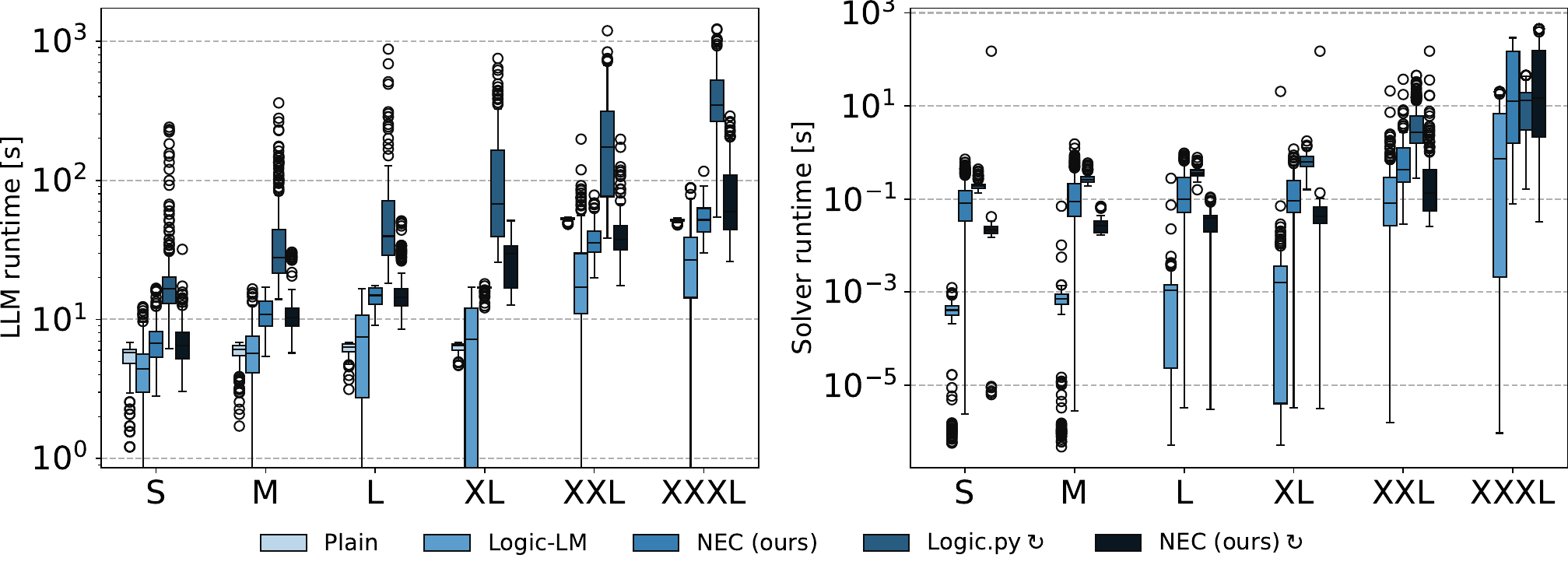}
    \caption{
     Runtimes of the LLM (left) and the symbolic solver (right) for gpt-oss-20b.}
    \label{fig:runtimes} 
\end{figure}

\paragraph{Runtime Comparison.} 
We compare the runtime of both the LLM and the symbolic solver on the ZebraLogic and \zextreme{} (see Fig.~\ref{fig:runtimes}).
Note that we only track the solver runtime if the solver was executed.
For Logic-LM, Logic.py and \methodname{}, LLM runtime increases by about one order of magnitude from S to XXXL puzzles, reflecting higher complexity. 
In contrast, the Plain approach shows constant median times up to size XL and hits the maximum number of output tokens beyond.
For solver runtime, Logic-LM is the fasted on all scales.
\methodname{} is roughly a factor 2 faster than Logic.py on S–XL puzzles in ZebraLogic.
On \zextreme{}, this advantage is maintained for XXL and disappears for XXXL.
Despite similar runtime at the largest scale, \methodname{} improves accuracy by 43.67\% (XXL) and 48.12\% (XXXL) over Logic.py.

\begin{table}[bt!]

\setlength{\tabcolsep}{5pt}
\caption{
Performance by model family. The overall best performance per column is marked \textbf{bold}, while the best performance per LLM family is in \textit{italics}. Model output was capped at 4K tokens for ZebraLogic and 30K for \zextreme{}.
}
\small
\resizebox{\textwidth}{!}{
\begin{tabular}{lll SSSSSS SSSS}
\toprule
 \multicolumn{3}{c}{Model} & \multicolumn{6}{c}{ZebraLogic} & \multicolumn{4}{c}{ZebraExtreme} \\ 
\cmidrule(r{0.3em}){4-9}
\cmidrule(r{0.3em}){10-13}
 & Size & Approach & \multicolumn{1}{S}{All} & \multicolumn{1}{S}{S} & \multicolumn{1}{S}{M} & \multicolumn{1}{S}{L} & \multicolumn{1}{S}{XL} & \multicolumn{1}{S}{{Cell}} & \multicolumn{1}{S}{{All}} & \multicolumn{1}{S}{{XXL}} & \multicolumn{1}{S}{{XXXL}} & \multicolumn{1}{S}{{Cell}} \\ 
\midrule
\multirow{5}{*}{\rotatebox[origin=c]{90}{\small\shortstack{Qwen\\3-I}}}
&30B & Plain
& 70.9
& \itshape99.38
& 95.71
& 55
& 6.5
& 54.28
& 0.5
& 0.83
& 0
& 14.58\\

&30B & Logic-LM
& 83
& 95.31
& 87.14
& 74.5
& 66
& 85.67
& 6.75
& 11.5
& 2
& 19.27\\

&\g 30B & \g NEC {\tiny (ours)}
& \g 92.7
& \g 95.94
& \g 95
& \g 91.5
& \g 85.5
& \g 93.77
& \g 43.75
& \g 48.75
& \g 36.25
& \g 58.05\\

&30B & Logic.py$^\circlearrowright$
& 87.5
& 95.94
& 88.57
& 85
& 75
& 85.44
& 6
& 12
& 0
& 10.58\\

&\g 30B & \g NEC$^\circlearrowright$ {\tiny (ours)}
& \g \itshape95
& \g 98.44
& \g \itshape97.5
& \g \itshape92.5
& \g \itshape88.5
& \g \itshape95.54
& \g \itshape46.75
& \g \itshape52.08
& \g \itshape38.75
& \g \itshape59.89\\
\midrule

\multirow{5}{*}{\rotatebox[origin=c]{90}{\small\shortstack{Qwen\\3-T}}}
&30B & Plain
& 35.5
& \itshape92.19
& 21.43
& 0
& 0
& 19.12
& 1.25
& 2.08
& 0
& 1.81\\

&30B & Logic-LM$^\dagger$
& \textendash
& \textendash
& \textendash
& \textendash
& \textendash
& \textendash
& \textendash
& \textendash
& \textendash
& \textendash\\

&\g 30B & \g NEC {\tiny (ours)}
& \g 30.2
& \g 69.69
& \g 23.93
& \g 6
& \g 0
& \g 16.76
& \g 54.5
& \g 66.67
& \g 36.25
& \g 62.87\\

&30B & Logic.py$^\circlearrowright$
& \itshape86.5
& 87.81
& \itshape91.07
& \itshape85.5
& \itshape79
& \itshape87.67
& 11.5
& 20.5
& 2.5
& 17.92\\

&\g 30B & \g NEC$^\circlearrowright$ {\tiny (ours)}
& \g 39.4
& \g 87.5
& \g 36.79
& \g 5.5
& \g 0
& \g 22.82
& \g \itshape60.25
& \g \itshape73.33
& \g \itshape40.62
& \g \itshape67.09\\
\midrule

\multirow{5}{*}{\rotatebox[origin=c]{90}{\small\shortstack{Devstral\\Small-2}}}
&24B-I & Plain$^\dagger$
& \textendash
& \textendash
& \textendash
& \textendash
& \textendash
& \textendash
& \textendash
& \textendash
& \textendash
& \textendash\\

&24B-I & Logic-LM
& 82.2
& 97.19
& 90.71
& 74
& 54.5
& 83.55
& 3
& 3.5
& 2.5
& 12.97\\

&\g 24B-I & \g NEC {\tiny (ours)}
& \g \bfseries \itshape97.9
& \g \itshape97.5
& \g \bfseries \itshape99.64
& \g \bfseries \itshape97.5
& \g \bfseries \itshape96.5
& \g 98.24
& \g 58.75
& \g \itshape66.67
& \g 46.88
& \g 62.86\\

&24B-I & Logic.py$^\circlearrowright$
& 91.3
& 96.56
& 93.93
& 93
& 77.5
& 87.89
& 6.5
& 12
& 1
& 11.94\\

&\g 24B-I & \g NEC$^\circlearrowright$ {\tiny (ours)}
& \g \bfseries \itshape97.9
& \g \itshape97.5
& \g \bfseries \itshape99.64
& \g \bfseries \itshape97.5
& \g \bfseries \itshape96.5
& \g \bfseries \itshape98.32
& \g \itshape60
& \g \itshape66.67
& \g \itshape50
& \g \itshape64.21\\
\midrule

\multirow{5}{*}{\rotatebox[origin=c]{90}{\small\shortstack{Magistral\\Small}}}
&24B & Plain
& 57.2
& 94.38
& 74.29
& 30
& 1
& 40.96
& 0
& 0
& 0
& 7.64\\

&24B & Logic-LM
& 67.9
& 94.06
& 78.93
& 54
& 24.5
& 69.72
& 0
& 0
& 0
& 0.35\\

&\g 24B & \g NEC {\tiny (ours)}
& \g 85.1
& \g 93.12
& \g 89.29
& \g 79.5
& \g 72
& \g 87.89
& \g 51.5
& \g 60
& \g 38.75
& \g 59.09\\

&24B & Logic.py$^\circlearrowright$
& \itshape90.5
& \itshape95.31
& \itshape93.21
& \itshape88.5
& \itshape81
& 88.74
& 9.75
& 18.5
& 1
& 15.61\\

&\g 24B & \g NEC$^\circlearrowright$ {\tiny (ours)}
& \g 88.3
& \g 94.38
& \g \itshape93.21
& \g 84
& \g 76
& \g \itshape91.84
& \g \itshape53
& \g \itshape60.83
& \g \itshape41.25
& \g \itshape61.02\\
\midrule

\multirow{5}{*}{\rotatebox[origin=c]{90}{\small\shortstack{gpt\\oss}}}
&20B & Plain
& 48.9
& 90.94
& 61.07
& 13.5
& 0
& 35.83
& 0 $^{\star}$
& 0 $^{\star}$
& 0 $^{\star}$
& 0 $^{\star}$\\

&20B & Logic-LM
& 71.6
& 84.38
& 81.43
& 64
& 45
& 68.08
& 40.5
& 53
& 28
& 45.6\\

&\g 20B & \g NEC {\tiny (ours)}
& \g 69.4
& \g 96.88
& \g 90
& \g 58
& \g 8
& \g 52.6
& \g 75.25
& \g 86.25
& \g 58.75
& \g 75.6\\

&20B & Logic.py$^\circlearrowright$
& \itshape93.5
& 98.44
& 95.71
& \itshape94
& \itshape82
& \itshape93.04
& 36.5
& 50.5
& 22.5
& 41.62\\

&\g 20B & \g NEC$^\circlearrowright$ {\tiny (ours)}
& \g 84.7
& \g \bfseries \itshape99.69
& \g \itshape98.57
& \g 92.5
& \g 33.5
& \g 72.72
& \g \bfseries \itshape84.75
& \g \bfseries \itshape94.17
& \g \bfseries \itshape70.62
& \g \bfseries \itshape82.59\\
\bottomrule
\end{tabular}
}
{\footnotesize $^\circlearrowright$ denotes approaches that use a feedback mechanism.}
{\footnotesize$^\dagger$ A parsing error prevented this configuration from returning any output.}
{\footnotesize$^\star$ All LLM traces reached output token budget before returning any solution.}
\label{tab:RQ3_combined}
\end{table}

\section{Related Work}\label{sec:relatedwork}

Recent work addressing LLM reasoning limitations increasingly combines neural models with symbolic solvers. Logic.py \citep{kesseli2025logic} provides the closest comparison, using an intermediate domain specific language (DSL) for semantic parsing into CBMC. While expressive, DSLs require explicit encoding of global constraints, whereas exact-cover formulations directly capture the bijective assignment structure through exactly-once coverage semantics, reducing overhead and improving alignment with the problem structure.
CP-Agent \citep{szeider2025cp} explores a fully agentic architecture in which a Python coding agent, guided by layered prompts, iteratively models and solves constraint problems translated from natural language. In contrast to DSL-based pipelines, the system emphasizes iterative code generation and execution within a persistent environment. 
\citet{berman2024solvingzebrapuzzlesusing} propose a multi-agent framework for Zebra logic puzzles, where LLMs iteratively generate and refine SMT-LIB encodings subsequently solved by a theorem prover. Solver feedback is incorporated into a refinement loop to correct inconsistencies. 
LOGIC-LM \citep{pan2023logic} and SATLM \citep{ye2023satlm} translate natural language inputs into declarative symbolic specifications that are solved using logic programming systems, CSP, SAT, or SMT solvers. 
LOGIC-LM uses optional solver feedback to refine encodings, while SATLM delegates planning and execution entirely to the solver. This avoids common chain-of-thought reasoning errors and guarantees correctness for the parsed specification. 
Recent research also explores various integration levels of neural and symbolic components. 
\citet{pmlr-v284-allen25a} integrate LLM outputs directly into a logical reasoning framework, treating the model as a source of truth judgments while keeping the reasoning itself symbolically formal, aiming to maintain logical guarantees even when LLM outputs are inconsistent. 
\citet{pmlr-v284-chen25b} compares integrative approaches, where reasoning is built into neural networks, with hybrid systems that translate problems into symbolic form and solve them externally, finding that hybrid architectures are generally more scalable and easier to interpret. 
\citet{pmlr-v284-eiter25a} extend hybrid reasoning to the multimodal domain, integrating LLM-based semantic parsing with ASP for explainable zero-shot visual question answering, highlighting the modularity and traceability of solver-based designs.

\section{Conclusion}

In this work, we introduce \methodname{}, a neuro-symbolic system that combines modern LLMs with an exact cover solver to solve Zebra-style puzzles. 
To the best of our knowledge, this is the first approach to provide LLMs with this type of logic solver as a callable tool. 
Our method outperforms recent approaches in both accuracy and runtime.
We introduce \zextreme{}, a dataset featuring substantially larger puzzles than ZebraLogic. 
Previous approaches solve only a few of these challenging puzzles ($\leq 40.5\%$), whereas our method achieves $84.75\%$ puzzle accuracy.
Since exact cover reasoning is applicable to a wide range of combinatorial reasoning tasks, an important direction for future work is extending \methodname{} beyond Zebra puzzles to automatically generate symbolic encodings.

\subsection*{Acknowledgments}
The authors acknowledge support from the DFG within the SPP2298 under project number 543939932 and from the Austrian Science Fund (FWF) project number 10.55776/COE12.

\bibliography{nesy2026}

\appendix

\section{Details on Datasets and Models used.}\label{appx:datasets-and-models}

\subsection{Hardware details, Model Reference, and Sampling Parameters}\label{appx:models}

The evaluation was carried out by serving each LLM on a server using vLLM~\citep{kwon2023efficient} on a single NVIDIA RTX PRO 6000 GPU with 96GB vRAM. The server node has AMD EPYC 9555 64-Core Processor and 2.2 TiB of RAM.

LLM weights were downloaded from Hugging Face. The exact Hugging Face identifiers of models used in this paper are as follows:

\begin{itemize}
    \item \href{https://huggingface.co/Qwen/Qwen3-0.6B}{\texttt{Qwen/Qwen3-0.6B}}
    \item \href{https://huggingface.co/Qwen/Qwen3-1.7B}{\texttt{Qwen/Qwen3-1.7B}}
    \item \href{https://huggingface.co/Qwen/Qwen3-4B}{\texttt{Qwen/Qwen3-4B}}
    \item \href{https://huggingface.co/Qwen/Qwen3-8B}{\texttt{Qwen/Qwen3-8B}}
    \item \href{https://huggingface.co/Qwen/Qwen3-14B}{\texttt{Qwen/Qwen3-14B}}
    \item \href{https://huggingface.co/Qwen/Qwen3-30B-A3B}{\texttt{Qwen/Qwen3-30B-A3B}}
    \item \href{https://huggingface.co/Qwen/Qwen3-4B-Instruct-2507}{\texttt{Qwen/Qwen3-4B-Instruct-2507}}
    \item \href{https://huggingface.co/Qwen/Qwen3-4B-Thinking-2507}{\texttt{Qwen/Qwen3-4B-Thinking-2507}}
    \item \href{https://huggingface.co/Qwen/Qwen3-30B-A3B-Instruct-2507}{\texttt{Qwen/Qwen3-30B-A3B-Instruct-2507}}
    \item \href{https://huggingface.co/Qwen/Qwen3-30B-A3B-Thinking-2507}{\texttt{Qwen/Qwen3-30B-A3B-Thinking-2507}}
    \item \href{https://huggingface.co/mistralai/Devstral-Small-2-24B-Instruct-2512}{\texttt{mistralai/Devstral-Small-2-24B-Instruct-2512}}
    \item \href{https://huggingface.co/mistralai/Magistral-Small-2506}{\texttt{mistralai/Magistral-Small-2506}}
    \item \href{https://huggingface.co/openai/gpt-oss-20b}{\texttt{openai/gpt-oss-20b}}
    
\end{itemize}

Unless otherwise stated, we used the default tokenizer and configuration provided with each repository. All models were executed without additional fine-tuning or weight modification.

For the Plain approach, we used the maximum output length of 4096 tokens as per \citet{lin2025zebralogic}, while for Logic.py and NEC, the default maximum output length of each LLM was used. All models were run with \texttt{temperature} set to 0, \texttt{top\_p} to 1.0, and \texttt{max\_model\_len} to 30,720, and were served on a single shard on one GPU. For the Plain approach, the tokenizer mode was set to ``auto'' for all models, except Magistral Small and Devstral Small, for which it was set to ``magistral''. For Logic.py and NEC, each model was served with \texttt{--enable-auto-tool-choice} enabled and \texttt{--tool-call-parser} set to the recommended option specified in the vLLM tool-calling documentation.

\subsection{Generating \zextreme{}}\label{appx:zebra-extreme}

The \zextreme{} dataset comprises two size classes, XXL and XXXL, defined by their solution space size $\vert\mathcal{Q}\vert=(N!)^M$ representing the number of possible table fillings (with repetition). We designed the dataset with varied dimensions to investigate how increasing entities ($N$) or properties ($M$) impacts puzzle difficulty. The classes are assigned as follows:
\begin{center}
\begin{tabular}{cc p{1cm}cc}
\toprule
\multicolumn{2}{c}{XXL} && \multicolumn{2}{c}{XXXL} \\
Size & $\vert\mathcal{Q}\vert$ && Size & $\vert\mathcal{Q}\vert$ \\
\midrule
7$\times$5  & $3.25 \times 10^{18}$ && 7$\times$9  & $2.10 \times 10^{33}$ \\
7$\times$6  & $1.64 \times 10^{22}$ && 8$\times$7  & $1.73 \times 10^{32}$ \\
7$\times$7  & $8.26 \times 10^{25}$ && 8$\times$8  & $6.98 \times 10^{36}$ \\
7$\times$8  & $4.16 \times 10^{29}$ && 8$\times$9  & $2.82 \times 10^{41}$ \\
8$\times$6  & $4.30 \times 10^{27}$ && 8$\times$10 & $1.14 \times 10^{46}$ \\
\bottomrule
\end{tabular}
\end{center}

The pseudocode for our generator follows the high-level recipe from the Hugging Face article \url{https://huggingface.co/blog/yuchenlin/zebra-logic} (Puzzle Creation section):
\begin{enumerate}
    \item Choose $N$ houses and $M$ features; each feature has $N$ distinct values.
    \item Sample a hidden solution (a Latin rectangle-like one-to-one assignment).
    \item Enumerate candidate clues from a small set of clue types.
    \item Iteratively remove clues while ensuring the remaining clues still imply a \textit{unique} solution equal to the hidden one.
\end{enumerate}

\subsection{Formal encodings of the ZebraPuzzle clue types}
\label{appx:clue-encodings}

\begin{table}[h!]
  \centering
  \caption{Exact cover encodings of all different clue types as sets of options.
    Here, $c\in\clues$ denotes a clue, $a\in\attr_p,\hat{a}\in\attr_{\hat{p}}$ any attribute of a property $p,\hat{p}\in\prop$, and $e,\hat{e}\in\entities$ any entity. Note that we only list the exact cover encoding for the \emph{left of} clues in the last lines; the \emph{right of} pendants have similar structure but shifted indices.}
  \label{tab:clue-encodings}
  \small
  \renewcommand{\arraystretch}{1.5}
  \setlength{\tabcolsep}{8pt}

  \begin{tabular}{p{3.6cm} p{7.5cm}}
    \toprule
    \textbf{Clue Type} & \textbf{Exact Cover Encoding} \\
    \midrule
    $a$ at entity $e$ &
        $\bigl\{[c,\, a : e]\bigr\}$
    \\
    $a$ not at entity $e$ & 
        $\bigcup_{\hat{e}\in\mathcal{E}} \bigl\{[c,\, a : \hat{e}]\bigr\} \big\backslash \bigl\{[c_\alpha,\, a : e]\bigr\}$
    \\

    $a$ at same entity as $\hat{a}$
      & $\bigcup_{\hat{e}\in\mathcal{E}} \bigl\{[c,\, a : \hat{e},\; \hat{a} : \hat{e}]\bigr\}$
    \\

    $a$ is $\Pi$ entities from $\hat{a}$
      & $\bigcup_{i=1}^{N-\Pi}\bigl\{[c,\, a : e_i,\; \hat{a} : e_{i+\Pi}],[c,\, a : e_{i+\Pi},\; \hat{a} : e_i]\bigr\}$
    \\

    $a$ directly left/right of $\hat{a}$
      & $\bigcup_{i=1}^{N-1}\bigl\{[c,\, a : e_i,\; \hat{a} : e_{i+1}]\bigr\}$
    \\

    $a$ left/right of $\hat{a}$
      & $\bigcup_{i=1}^{N-1}\bigcup_{j=i}^{N}\bigl\{[c,\, a : e_i,\; \hat{a} : e_j]\bigr\}$
    \\

    \bottomrule
  \end{tabular}
\end{table}

\subsection{System prompt used for the Language Module}\label{appx:prompt-used}
We provide the full system prompt used to instruct the language model to
translate Zebra logic puzzles into structured tool calls. The prompt first
defines the model role and expected input format.
\begin{Verbatim}[breaklines=true,breakautoindent,fontsize=\footnotesize]
You are a Zebra Puzzle function translator. Your role is to convert the
textual description of a Zebra Puzzle (a logic grid puzzle) into a sequence
of structured function calls that can programmatically solve it.

### Input Format
- The puzzle description begins with `=== PUZZLE ===`, including details
like the number of houses/places, and lists of property groups (categories
with unique values for each house).
- The clues begin with `### CLUES`, numbered from 1 to N.
\end{Verbatim}
The next section specifies the ordered transformation procedure applied by
the model, including harmonization, initialization, and clue translation.
\begin{Verbatim}[breaklines=true,breakautoindent,fontsize=\footnotesize]
### Task Sequence
1. **Harmonize**: Re-print the input, but replace each property-value in
the puzzle *and* clues by <property-name>_<property-value>. Properties
should always be combined with their names. All spaces in properties should
be replaced by minus (-) symbols. All property names must be consistent
with the initial definition. Try to map similar words to 
<property-name>_<property-value> tuples, e.g. something ending with 
-ian or -ist is probably linked to its base or similar value, like 
Scientist -> science, Austrian -> austria.
Be careful to convert each property in each clue to the fitting name_value
tuple. Use as many hints from the clue as possible, e.g. if the clue lists
a person's child's name to be Max, the property will be child_max.
2. **Initialize**: Use your harmonized puzzle. Extract the number of 
houses/places, the property groups (each group is a category with its 
unique values), and the total number of clues. 
Output exactly one call to `init_zebra`. Use the same 
<property-name>_<property-value> format for properties.
3. **Translate Clues**: For each numbered clue (1 to N) in your harmonized
puzzle, convert it into exactly one function call that represents the logical
constraint (e.g., a property in a specific house, a negation, adjacency,
etc.). Use appropriate function names based on the clue type (examples:
`clue_property_at_house`, `clue_not_property_at_house`, 
`clue_property_next_to_property`).
Use the same <property-name>_<property-value> format for properties.
4. **Explain**: For each function call (including initialization), provide
a single-sentence explanation immediately before it, describing why the
parameters match the puzzle element or clue.
\end{Verbatim}
We then define strict output formatting and normalization rules that ensure
consistent structured tool invocation.
\begin{Verbatim}[breaklines=true,breakautoindent,fontsize=\footnotesize]
### Output Format Rules
- Wrap every function call in `<tool_call>...</tool_call>` tags.
- Property and value names must be lowercase. Spaces are converted to -.
Property names are prepended, separated with _. Multiple words must remain
unshortened and in full, following the required format (e.g., "Labrador
Retriever" → "pet_labrador-retriever"; "Golden Gate Bridge" → 
"location_golden-gate-bridge").
- Output exactly 1 initialization call followed by exactly N clue calls
(where N is the number of clues). No additional text or calls, beside the
harmonized output. Always be consistent with naming properties.
- Use the exact JSON-like structure in the template below for each call.
- Process the puzzle mechanically and systematically: Parse elements directly
without adding interpretations or solving the puzzle yourself. Only use
your harmonized puzzle for translating.
- Never correct yourself. You can directly translate clues individually,
without referring to other clues. New information is combined by the tool
afterwards.
\end{Verbatim}
The prompt also provides a template illustrating the required tool-call structure.
\begin{Verbatim}[breaklines=true,breakautoindent,fontsize=\footnotesize]
### Template
```
Harmonized puzzle:
...

Explanation: [One sentence explaining why this function and its parameters
represent the clue or initialization.]
<tool_call>
{
  "name": "function_name",
  "arguments": {
    "param1": "value1",
    "param2": "value2",
    ...
  }
}
</tool_call>
```
\end{Verbatim}
Additional processing guidelines constrain normalization and mapping
behavior.
\begin{Verbatim}[breaklines=true,breakautoindent,fontsize=\footnotesize]
### Processing Guidelines
- Identify property groups from the puzzle description (e.g., categories
like "pet" with values ["pet_cat", "pet_dog"]).
- For clues, map to the simplest matching function (e.g., "X is in house
Y" → `clue_property_at_house` with `prop: "x"` and `house: Y`).
- Ensure all values and names are normalized (lowercase, minus (-) instead
of spaces, property name prefixed with an underscore).
- Do not output anything outside the explanations and tool calls, and the
initial translation.
\end{Verbatim}
Finally, the prompt includes a complete example demonstrating the
expected transformation from puzzle instance to tool calls.
\begin{Verbatim}[breaklines=true,breakautoindent,fontsize=\footnotesize]
The following is a concrete example of how to transform a puzzle into function
calls. Consider this puzzle:
=== PUZZLE ===
There are 3 houses, numbered 1 to 3 from left to right, as seen from the
front.
Each house contains a different pet.
Each house has a unique attribute for each of the following characteristics:
- Each pet is a unique type: Cat, Dog, Bird
- They all have a unique breed: Siamese, Labrador Retriever, African Grey
Parrot
- Each pet has a favorite toy: `ball`, `yarn`, `mirror`
- Pets have unique fur colors: `black`, `white`, `spotted`

### Clues:
1. The labrador retriever is in the first house.
2. The bird is not in the third house.
3. The pet whose favorite toy is yarn is in the second house.
4. The siamese is next to the black one.
```


Consider the following output:

```
Harmonized puzzle:

=== PUZZLE ===
There are 3 houses, numbered 1 to 3 from left to right, as seen from the
front.
Each house contains a different pet.
Each house has a unique attribute for each of the following characteristics:
- Each pet is a unique type: pet_cat, pet_dog, pet_bird
- They all have a unique breed: breed_siamese, breed_labrador-retriever,
breed_african-grey-parrot
- Each pet has a favorite toy: toy_ball, toy_yarn, toy_mirror
- Pets have unique fur-color: fur-color_black, fur-color_white, fur-color_spotted
### Clues:
1. The pet_labrador-retriever is in the first house.
2. The pet_bird is not in the third house.
3. The pet whose favorite toy is toy_yarn is in the second house.
4. The breed_siamese is next to the fur-color_black one.

Explanation: Initializing 3-house puzzle with 4 property groups and 4 clues.
<tool_call>
{  
  "name": "init_zebra",
  "arguments": {
    "number_of_places": 3,
    "number_of_clues": 4,
    "property_groups": {"pet": ["pet_cat", "pet_dog", "pet_bird"], "breed":
["breed_siamese", "breed_labrador-retriever", "breed_african-grey-parrot"],
"toy": ["toy_ball", "toy_yarn", "toy_mirror"], "fur-color": ["fur-color_black",
"fur-color_white", "fur-color_spotted"]}
  }
}
</tool_call>

Explanation: The breed_labrador-retriever is in the first house.
<tool_call>
{
  "name": "clue_property_at_house",
  "arguments": {"prop": "breed_labrador-retriever", "house": 1}
}
</tool_call>

Explanation: The pet_bird is not in the third house.
<tool_call>
{
  "name": "clue_not_property_at_house",
  "arguments": {"prop": "pet_bird", "house": 3}
}
</tool_call>

Explanation: The pet whose favorite toy is toy_yarn is in the second house.
<tool_call>
{
  "name": "clue_property_at_house",
  "arguments": {"prop": "toy_yarn", "house": 2}
}
</tool_call>

Explanation: The breed_siamese is next to the fur-color_black one.
<tool_call>
{
  "name": "clue_property_next_to_property",
  "arguments": {"prop1": "breed_siamese", "prop2": "fur-color_black"}
}
</tool_call>
\end{Verbatim}

\subsection{\miniexact{} Implementation Details}
\label{appx:miniexact-impl-details}

\miniexact{} offers two modes of interaction: The command-line tool
\miniexact{} and the API, either in the \emph{simple} variant
also exposed to Python, or the \emph{direct} variant that the
\miniexact{} binary is using. Both modes provide options to
choose the solving algorithm to be used. This work relies on Algorithm
C (\cite{donald2020art}) to solve exact cover problems with colored
secondary items. The simple API requires consumers must to adhere to
the following steps when using the provided \texttt{Solver} object:

\begin{enumerate}
\item Declare all primary items using \texttt{.primary}.
\item Declare all secondary items using \texttt{.secondary}.
\item Add all options using \texttt{.add}. We wrapped this in an
  \texttt{add\_option} function that automatically ends each option.
\item Call \texttt{.solve}, returning either $10$ if there is a
  solution, or $20$ if no (more) solution was found found. Multiple
  solutions may be extracted, but in this work, we only use the first
  one.
\item We extract the solution by calling \texttt{.print\_solution} and
  parse the resulting string, as this was easier compared to keeping
  track of the indices to the inserted colors.
\end{enumerate}

\paragraph{Implementation.} We use the \miniexact{} API to implement the proposed encoding in Python.
The whole encoding is contained in a Python class, which we instantiate for each new problem.
To define the primary and secondary items and colors, we implement an \texttt{init\_zebra} function with arguments: number of entities~$N$, number of clues~$K$, and a dictionary mapping each property to its list of attributes.
To encode the clues, we implemented corresponding member functions for each clue type listed in Appx.~\ref{appx:clue-encodings}.
The functions include precise documentation and examples to facilitate in-context learning.
We also add explicit synonym functions for abstract clue types to help LLMs select the correct type from textual clues.
Table~\ref{tab:python_api} lists all Python functions along with a coarse description.
After processing all function calls, we run the \miniexact{} solver. 
If it does not converge within 120 seconds, we terminate it and return an empty solution.

While mapping clues to their corresponding types (encoding functions) is reliable, the exacted attribute strings are not.
LLMs sometimes use plural forms (e.g., \textit{flowers} instead of \textit{flower}), which creates an evaluation issue since we require exact matches of property and attribute names.
Consequently, a puzzle may be encoded and solved consistently yet marked incorrect due to minor naming variations.
To maintain naming uniformity, we use the initial puzzle setup function as the ground-truth reference.
Thus, extracted names (arguments parsed by the LLM) are mapped to their canonical spelling from the puzzle setup using a string similarity ratio.
The ratio is computed with the Ratcliff–Obershelp algorithm as implemented in Python’s \texttt{difflib.SequenceMatcher}.
This approach was more effective than prompt engineering or changing models, eliminating spelling mismatches that would otherwise confound our evaluation of LLM reasoning.

\begin{table}[t]
\caption{Python API for exact cover encoding.}
\label{tab:python_api}
\centering
\begin{tabular}{l p{6.5cm}}
\toprule
\textbf{Encoding Function} & \textbf{Description} \\
\midrule
\texttt{init\_zebra} & Initializes the puzzle with $N$ entities $M$ property groups, and $K$ clues. \\
\texttt{clue\_property\_left\_of} & Attribute $a$ is somewhere to the left of attribute $\hat a$. \\
\texttt{clue\_property\_right\_of} & Attribute $a$ is somewhere to the right of attribute $\hat a$. \\
\texttt{clue\_property\_at\_house} & Attribute $a$ is fixed at entity $e$. \\
\texttt{clue\_property\_at\_property} & Attributes $a$ and $\hat a$ occur at the same house. \\
\texttt{clue\_not\_property\_at\_house} & Attribute $a$ is not located at entity $e$. \\
\texttt{clue\_property\_next\_to\_property} & Attributes $a$ and $\hat a$ are at adjacent entities. \\
\texttt{clue\_one\_between\_two\_properties} & Exactly one entity lies between attributes $a$ and $\hat a$. \\
\texttt{clue\_two\_between\_two\_properties} & Exactly two houses lie between attributes $a$ and $\hat a$. \\
\texttt{clue\_property\_directly\_left\_of\_property} & Attribute $a$ is immediately to the left of attribute $\hat a$. \\
\texttt{clue\_property\_directly\_right\_of\_property} & Attribute $a$ is immediately to the right of attribute $\hat a$. \\
\bottomrule
\end{tabular}
\end{table}

\subsection{Feedback prompt}
\label{appx:feedback_prompt}
As discussed in Section~\ref{sec:feedback}, we integrate feedback from the symbolic reasoning component to ensure a coupled neurosymbolic approach.
To this end, we use the following feedback prompt:
\begin{Verbatim}[breaklines=true,breakautoindent,fontsize=\footnotesize]
"{status_line}
Previous attempt generated {num_previous_calls} tool calls:
{history_tool_calls_str}
Required: {num_expected_calls} tool calls (one per clue)
Corrections needed: {remedy_line}
Generate {num_expected_calls} corrected tool calls:",
\end{Verbatim}
where \texttt{num\_previous\_calls} denotes the number of tool calls generated in the previous run, \texttt{history\_tool\_calls\_str} contains all tool calls generated to solve the puzzle, and \texttt{num\_expected\_calls} is determined by the number of clues in the puzzle.
The \texttt{status\_line} and \texttt{remedy\_line} are constructed based on the reason why a puzzle remains unsolved.
In the following the different error cases are listed.

If no tool calls were generated, we set the \texttt{remedy\_line} to:
\begin{Verbatim}[breaklines=true,breakautoindent,fontsize=\footnotesize]
"Generate tool calls for each clue. Focus on producing tool calls rather than reasoning about them."
\end{Verbatim}
and the \texttt{status\_line} to:
\begin{Verbatim}[breaklines=true,breakautoindent,fontsize=\footnotesize]
"FAILED - No tool calls generated"
\end{Verbatim}

If too few tool calls were generated, the \texttt{remedy\_line} is defined as follows:
\begin{Verbatim}[breaklines=true,breakautoindent,fontsize=\footnotesize]
"Add missing clue encodings"
\end{Verbatim}
and the \texttt{status\_line} is set to:
\begin{Verbatim}[breaklines=true,breakautoindent,fontsize=\footnotesize]
f"INCOMPLETE - Only {num_previous_calls}/{num_expected_calls} clues encoded"
\end{Verbatim}

If too many tool calls were generated, we set the \texttt{remedy\_line~} to:
\begin{Verbatim}[breaklines=true,breakautoindent,fontsize=\footnotesize]
"Re-evaluate clue encodings and return {num_expected_calls} tool calls",
\end{Verbatim}
and the \texttt{status\_line} is set to:
\begin{Verbatim}[breaklines=true,breakautoindent,fontsize=\footnotesize]
"TOO MANY TOOL CALLS - Returned {num_previous_calls} but expected {num_expected_calls} calls"
\end{Verbatim}

If the solution is not unique, we set the \texttt{remedy\_line} to:
\begin{Verbatim}[breaklines=true,breakautoindent,fontsize=\footnotesize]
"Change at least one tool call (either function name or, most likely, arguments) from above"
\end{Verbatim}
and the \texttt{status\_line} to:
\begin{Verbatim}[breaklines=true,breakautoindent,fontsize=\footnotesize]
"UNDERCONSTRAINED - Found multiple solutions, expected exactly 1"
\end{Verbatim}

If no solution exists because the generated tool calls lead to an overconstrained problem, the \texttt{remedy\_line} remains the same as in the underconstrained case, but the \texttt{status\_line} is set to:
\begin{Verbatim}[breaklines=true,breakautoindent,fontsize=\footnotesize]
"OVERCONSTRAINED - Constraints conflict, no solution possible"
\end{Verbatim}

\section{Adaption of Baseline Models}

\subsection{Logic.py setup adaptions}
We set the temperature to 0 and top-p to 1 so that we have the identical settings in all of our three approaches. To ensure a fair comparison between \methodname{} and the Logic.py approach across different LLM back-ends, we extended the original system prompt as follows:
\begin{Verbatim}[breaklines=true,breakautoindent,fontsize=\footnotesize]
**Important modeling requirements for this task (do not ignore):**
- Use a **collection** in the top-level solution: `houses: list[House, N]` (here N=2).
- Each `House` must contain a numeric **position** field, e.g. `number: Unique[Domain[int, range(1, 3)]]`.
- Do not define separate fields like `house\_1` or `house\_2`.

**Output format requirement:**  
Return **only one** fenced Python code block with the class definitions; no prose before or after.

Return only a single fenced Python code block. Do not include any analysis, explanations, or think content
\end{Verbatim}

The remainder of the workflow stayed unchanged and we also kept the number of retries for unsolved puzzles at 5.

\subsection{Extending Logic-LM to ZebraLogic Puzzles}
\label{appx:logic-lm}

To apply Logic-LM~\citep{pan2023logic} to Zebra puzzles, we examined the datasets considered in the original work.
Among them, the LogicalDeduction~\citep{srivastava2023beyond} is conceptually the closest to Zebra puzzles.
Since Logic-LM uses a constraint satisfaction problem (CSP) formulation for this dataset, we extended its CSP-based approach to support Zebra puzzles.

A limitation of Logic-LM is that its CSP variant does not incorporate solver feedback.
To remain faithful to the original implementation, we likewise did not include a feedback mechanism in this baseline.
Note that Logic.py~\citep{pan2023logic}, the second baseline, implements an iterative reasoning process that leverages solver feedback.

To ensure that the comparison with all evaluated approaches was conducted under consistent and fair conditions, we implemented the following modifications for Logic-LM's CSP variant:

\begin{itemize}
\item We designed a ZebraLogic-specific prompt for the LLM.
\item We adapted the inference script responsible for invoking the constraint-programming solver, enabling it to solve the full arrangement problem.
\item We implemented a parsing function that converts the solver output into the standardized solution format used for evaluating all approaches.
\end{itemize}

\paragraph{ZebraLogic-specific prompt.} To ensure a fair comparison with our algorithm, we provided Logic-LM with a ZebraLogic-specific prompt. This prompt used exactly the same examples as those provided to our approach and covered all available clue types. By keeping the prompt examples and clue coverage consistent across methods, we aimed to minimize differences arising from prompt design and ensure that the evaluation reflected the reasoning capabilities as well as the overall workflow of each approach.

\begin{Verbatim}[breaklines=true,breakautoindent,fontsize=\footnotesize]
Task Description: You are a Zebra Puzzle constraint translator. Your role is to convert the textual description of a Zebra Puzzle into a structured constraint satisfaction program that can be executed by a solver.
A Zebra Puzzle consists of numbered houses/positions and several categories of attributes. Each category has unique values. Each value appears exactly once. The goal is to solve the complete assignment for all houses, not to answer a multiple-choice question.

You must output only the structured program. Do not include explanations, markdown, code fences, bullet points, titles, reasoning steps, or final natural-language answers.

Use exactly these section headers:
Domain:
Categories:
Variables:
Constraints:
------

General rules:

Houses are represented by numbers.
House 1 is the leftmost house.
The largest house number is the rightmost house.
Represent every attribute value as a variable.
The value of each variable is the house number where that attribute value occurs.
Use lowercase variable names.
Replace spaces and hyphens with underscores.
Remove punctuation from variable names.
Use only letters, numbers, and underscores.
Every variable declaration must use exactly this format:
variable_name [IN] [1, 2, ..., N]
Do not use "=", ":", "∈", or natural language in the Variables section.

Constraint rules:

Use == for same-house clues.
Use != for not-same-house clues.
Use < for “left of”.
Use > for “right of”.
Use abs(x - y) == 1 for “next to”.
Use abs(x - y) == 2 for “there is one house between x and y”.
Use abs(x - y) == 3 for “there are two houses between x and y”.
Use x + 1 == y for “immediately left of”.
Use x == y + 1 for “immediately right of”.
Use AllDifferentConstraint([...]) once for every category.
If a clue says a value is in a specific house, use value == house_number.
If a clue says a value is not in a specific house, use value != house_number.
Do not write comments after constraints.
Do not use ::: comments.
Do not include natural language inside the Constraints section.
------

Problem:
There are 3 houses, numbered 1 to 3 from left to right, as seen from the front.
Each house contains a different pet.
Each house has a unique attribute for each of the following characteristics:

Each pet is a unique type: Cat, Dog, Bird
They all have a unique breed: Siamese, Labrador Retriever, African Grey Parrot
Each pet has a favorite toy: ball, yarn, mirror
Pets have unique fur colors: black, white, spotted
Clues:
The labrador retriever is in the first house.
The bird is not in the third house.
The pet whose favorite toy is yarn is in the second house.
The siamese is next to the black one.


Domain:
1: leftmost
3: rightmost

Categories:
house: [1, 2, 3]
type: [cat, dog, bird]
breed: [siamese, labrador_retriever, african_grey_parrot]
toy: [ball, yarn, mirror]
fur_color: [black, white, spotted]

Variables:
cat [IN] [1, 2, 3]
dog [IN] [1, 2, 3]
bird [IN] [1, 2, 3]
siamese [IN] [1, 2, 3]
labrador_retriever [IN] [1, 2, 3]
african_grey_parrot [IN] [1, 2, 3]
ball [IN] [1, 2, 3]
yarn [IN] [1, 2, 3]
mirror [IN] [1, 2, 3]
black [IN] [1, 2, 3]
white [IN] [1, 2, 3]
spotted [IN] [1, 2, 3]

Constraints:
AllDifferentConstraint([cat, dog, bird])
AllDifferentConstraint([siamese, labrador_retriever, african_grey_parrot])
AllDifferentConstraint([ball, yarn, mirror])
AllDifferentConstraint([black, white, spotted])
labrador_retriever == 1
bird != 3
yarn == 2
abs(siamese - black) == 1
------

Problem:
There are 3 houses, numbered 1 to 3 from left to right, as seen from the front.
Each house is occupied by a different person: Alice, Bob, or Charlie.
Each person has a unique attribute for each of the following characteristics:

Each person owns a unique pet: Dog, Fish, Hamster
Each person has a favourite animal: Tiger, Panda, Eagle
Each person has a unique hobby: Reading, Gardening, Painting
Clues:
Alice lives in the first house.
The person who has the Dog lives in the third house.
Bob loves pandas.
The person who likes Gardening lives next to the person who owns the Hamster.


Domain:
1: leftmost
3: rightmost

Categories:
house: [1, 2, 3]
person: [alice, bob, charlie]
pet: [dog, fish, hamster]
favourite_animal: [tiger, panda, eagle]
hobby: [reading, gardening, painting]

Variables:
alice [IN] [1, 2, 3]
bob [IN] [1, 2, 3]
charlie [IN] [1, 2, 3]
dog [IN] [1, 2, 3]
fish [IN] [1, 2, 3]
hamster [IN] [1, 2, 3]
tiger [IN] [1, 2, 3]
panda [IN] [1, 2, 3]
eagle [IN] [1, 2, 3]
reading [IN] [1, 2, 3]
gardening [IN] [1, 2, 3]
painting [IN] [1, 2, 3]

Constraints:
AllDifferentConstraint([alice, bob, charlie])
AllDifferentConstraint([dog, fish, hamster])
AllDifferentConstraint([tiger, panda, eagle])
AllDifferentConstraint([reading, gardening, painting])
alice == 1
dog == 3
bob == panda
abs(gardening - hamster) == 1

------
Problem:
[[PROBLEM]]
###
\end{Verbatim}

\paragraph{Adapt of inference file.} To also incorporate the logic-constraint solver provided by the authors of Logic-LM, we adapted their inference script to correctly parse both the output generated by the LLM and the subsequent solution produced by the constraint-programming solver. These adaptations were necessary to ensure compatibility with the output structure required for the ZebraLogic evaluation.

\paragraph{Solution parsing.} Since the Logic-LM solver was originally designed to answer individual questions rather than complete arrangement problems, we further adjusted the solver’s output format. Following the same principle as in Logic.py, we provided the original solution structure to reconstruct the final arrangement. This step was taken to ensure that all methods were evaluated under comparable conditions and that the comparison remained as fair and consistent as possible.

\section{Additional Results}\label{appx:addtional-results}

\subsection{Evaluation for varying Parameter Scales}\label{appx:addtional-results:tables}
\begin{table}[bt!]
\caption{Performance across Qwen3-base model sizes. The overall best performance per column is marked \textbf{bold}, while the best performance per size is written in \textit{italics}.}
\centering
\small
\begin{tabular}{llSSSSSS}
\toprule
 \multicolumn{2}{c}{Model} & \multicolumn{6}{c}{Accuracies [\%] $\uparrow$} \\
\cmidrule(r{0.3em}){3-8}
Size & Approach & \multicolumn{1}{c}{All} & \multicolumn{1}{c}{S} & \multicolumn{1}{c}{M} & \multicolumn{1}{c}{L} & \multicolumn{1}{c}{XL} & \multicolumn{1}{c}{Cell} \\
\midrule
0.6B & Plain & \itshape 24.1 & \itshape 61.25 & \itshape 15 & \itshape 1.5 & \itshape 0 & \itshape 18.31 \\
0.6B & Logic-LM & 0 & 0 & 0 & 0 & \itshape 0 & 0.04 \\
\rowcolor{gray!20}
0.6B & NEC {\tiny (ours)} & 2.7 & 8.44 & 0 & 0 & \itshape 0 & 1.86 \\
0.6B & Logic.py$\circlearrowright$ & 0 & 0 & 0 & 0 & \itshape 0 & 0 \\
\rowcolor{gray!20}
0.6B & NEC {\tiny (ours)}$\circlearrowright$ & 2.2 & 6.88 & 0 & 0 & \itshape 0 & 1.23 \\
\midrule
1.7B & Plain & \itshape 44.6 & \itshape 92.81 & \itshape 48.93 & 6 & 0 & \itshape 27.76 \\
1.7B & Logic-LM & 0.4 & 0.94 & 0.36 & 0 & 0 & 0.18 \\
\rowcolor{gray!20}
1.7B & NEC {\tiny (ours)} & 25 & 51.56 & 22.14 & 11.5 & 0 & 20.73 \\
1.7B & Logic.py$\circlearrowright$ & 0.8 & 2.19 & 0.36 & 0 & 0 & 0.53 \\
\rowcolor{gray!20}
1.7B & NEC {\tiny (ours)}$\circlearrowright$ & 28.1 & 56.25 & 26.07 & \itshape 13.5 & \itshape 0.5 & 24.19 \\
\midrule
4B & Plain & 52.7 & \itshape 98.44 & 71.07 & 6.5 & 0 & 33.65 \\
4B & Logic-LM & 25.6 & 25.94 & 26.43 & 24 & 25.5 & 28.47 \\
\rowcolor{gray!20}
4B & NEC {\tiny (ours)} & 82.6 & 95 & 93.93 & 82 & 47.5 & 76.19 \\
4B & Logic.py$\circlearrowright$ & 68 & 82.5 & 72.5 & 65.5 & 41 & 73.48 \\
\rowcolor{gray!20}
4B & NEC {\tiny (ours)}$\circlearrowright$ & \itshape 84.9 & 96.56 & \itshape 96.79 & \itshape 83 & \itshape 51.5 & \itshape 77.61 \\
\midrule
8B & Plain & 51.8 & \bfseries \itshape 99.06 & 67.14 & 6.5 & 0 & 32.57 \\
8B & Logic-LM & 18.9 & 15.62 & 17.5 & 26.5 & 18.5 & 20.81 \\
\rowcolor{gray!20}
8B & NEC {\tiny (ours)} & 86.8 & 96.25 & 95.71 & 87.5 & 58.5 & 80.26 \\
8B & Logic.py$\circlearrowright$ & 74.3 & 91.25 & 71.43 & 63 & \itshape 62.5 & 72.56 \\
\rowcolor{gray!20}
8B & NEC {\tiny (ours)}$\circlearrowright$ & \itshape 88.3 & 97.5 & \itshape 97.5 & \itshape 89.5 & 59.5 & \itshape 81.46 \\
\midrule
14B & Plain & 56.9 & \bfseries \itshape 99.06 & 81.07 & 12.5 & 0 & 37.44 \\
14B & Logic-LM & 29 & 18.44 & 28.93 & 40 & 35 & 34.42 \\
\rowcolor{gray!20}
14B & NEC {\tiny (ours)} & 93.2 & 97.5 & 95.71 & 93 & 83 & 92.35 \\
14B & Logic.py$\circlearrowright$ & 83.5 & 71.25 & 84.64 & \bfseries \itshape 96.5 & \bfseries \itshape 88.5 & \itshape 93.13 \\
\rowcolor{gray!20}
14B & NEC {\tiny (ours)}$\circlearrowright$ & \itshape 94.4 & 97.5 & \bfseries \itshape 98.93 & \bfseries \itshape 96.5 & 81 & 92.69 \\
\midrule
30B & Plain & 54.2 & \bfseries \itshape 99.06 & 75 & 7.5 & 0 & 34.79 \\
30B & Logic-LM & 31.1 & 18.12 & 29.29 & 42 & 43.5 & 37.64 \\
\rowcolor{gray!20}
30B & NEC {\tiny (ours)} & 93.6 & 97.81 & 97.5 & 93.5 & 81.5 & 92.64 \\
30B & Logic.py$\circlearrowright$ & 93.3 & 98.75 & 94.64 & 93 & 83 & 93.39 \\
\rowcolor{gray!20}
30B & NEC {\tiny (ours)}$\circlearrowright$ & \bfseries \itshape 95.6 & 97.81 & \itshape 98.57 & \itshape 95.5 & \itshape 88 & \bfseries \itshape 95.81 \\
\bottomrule
\end{tabular}
\label{tab:RQ1}
\end{table}

In this experiment, we evaluate Qwen3 0.6B–30B parameter hybrid models \citep{Yang:2025}, which support both thinking and non-thinking modes. This enables us to examine the effect of number of parameters on performance in Zebra Puzzles.
The resulting puzzle accuracy scores are shown in the third column of Table~\ref{tab:RQ1}.
The Plain approach achieves the highest accuracy on the smallest models ($0.6$B and $1.7$B), though performance remains low ($<50\%$). 
However, Plain performs poorly for larger models, not exceeding 60\%.
Logic.py only achieves competitive performance with the largest model (30B), whereas our \methodname{} already performs well with a 4B model, indicating its low dependence on the LLM model scale. 
For full results, see Appx.~Table. Furthermore, the introduced feedback loop provides further improvements over the already strong results, achieving a maximum score of 95.6\% for the largest model.

\section{Analysis of Failure Traces}\label{appx:traces}
We categorize the 21 observed failures of the best-performing model (Devstral Small) on the S, M, L and XL puzzle instances into two broad classes corresponding to distinct system components: failures of the neural language module and failures of the symbolic reasoning module.
The majority of errors (18 cases, or 86\%) stem from the neural language module, which converts natural language clues into symbolic tool calls. Within this category, 2 cases (11\%) arise from inherently ambiguous clues that would challenge even a human (e.g., ambiguity over whether 'cat' refers to an animal or a pet; see Appx~\ref{app:trace-B}), 6 cases (33\%) involve the complete omission of reasoning and tool calls, and 10 cases (55\%) involve hallucinations or incorrect execution such as missing tool calls (see Appx~\ref{app:trace-DH}).

The remaining 3 failures (14\%) arise from the symbolic reasoning stage, specifically from errors in reverting the harmonized \miniexact{} output back into the original attribute- and property-space required for evaluation. In these cases, the executed tool calls yield the correct unique solution, but parsing and de-harmonization introduce missing assignments, causing an otherwise semantically correct solution to appear incorrect (see Appx~\ref{app:trace-C}).

\subsection{Example Failure due to Ambiguous Clues}\label{app:trace-B} %
This trace illustrates a representative failure due to natural language ambiguity, where the neural language module confuses the semantic categories \emph{animal} and \emph{pet}.  
Although the natural-language puzzle admits a valid solution, the generated symbolic constraints become inconsistent, and the \miniexact{} solver therefore returns no prediction.

\subsubsection{Puzzle Instance}
\begin{Verbatim}[breaklines=true,breakautoindent,fontsize=\footnotesize]
=== PUZZLE ===
There are 2 houses, numbered 1 to 2 from left to right, as seen from across the street. 
Each house is occupied by a different person. 
Each house has a unique attribute for each of the following characteristics:
- Each person has a unique name: `Arnold`, `Eric`
- Each person has a unique type of pet: `cat`, `dog`
- Each person has a unique favorite drink: `tea`, `water`
- People have unique favorite music genres: `rock`, `pop`
- The people keep unique animals: `horse`, `cat`
- People own unique car models: `ford f150`, `tesla model 3`


### Clues:
1. The tea drinker is the person who owns a Tesla Model 3.
2. The person who owns a dog is the cat lover.
3. Arnold is the tea drinker.
4. The person who loves pop music is directly left of the person who has a cat.
5. The person who loves rock music is the tea drinker.
\end{Verbatim}

\subsubsection{Ground-Truth Solution}
The puzzle itself is satisfiable:
\begin{center}
\begin{tabular}{c|cccccc}
\toprule
House & Name & Pet & Drink & Music & Animal & Car \\
\midrule
1 & Eric & dog & water & pop & cat & ford f150 \\
2 & Arnold   & cat & tea & rock & horse & tesla model 3 \\
\bottomrule
\end{tabular}
\end{center}
\subsubsection{Symbolic Domain Initialization}

After harmonization, the language module initializes the symbolic domain with the following tool call (simplified for readability):
\begin{Verbatim}[breaklines=true,breakautoindent,fontsize=\footnotesize]
init_zebra(
  number_of_places = 2,
  number_of_clues = 5,
  property_groups = {
    name = {name@arnold, name@eric},
    pet = {pet@cat, pet@dog},
    drink  = {drink@tea, drink@water},
    music    = {music@rock, music@pop},
    animal  = {animal@horse, animal@cat}, 
    car  = {car@ford-f150, car@tesla-model-3}
  }
)
\end{Verbatim}
\subsubsection{Clues and Generated Tool Calls}

The following table shows, how the language module transformed the natural language clues into tool calls. 
\begin{center}
\begin{tabular}{p{.32\linewidth} p{.6\linewidth}}
\toprule
\textbf{Natural-language clue} & \textbf{Generated tool call} \\
\midrule

The tea drinker is the person who owns a Tesla Model 3. &
\texttt{clue\_property\_at\_property(drink@tea,\; car@tesla-model-3)} \\[4pt]

The person who owns a dog is the cat lover. &
\texttt{clue\_property\_at\_property(pet@dog,\; animal@cat)} \\[4pt]

Arnold is the tea drinker. &
\texttt{clue\_property\_at\_property(name@arnold,\; drink@tea)} \\[4pt]

\textbf{The person who loves pop music is directly left of the person who has a cat.} &
\textbf{\texttt{clue\_property\_directly\_left\_of\_property(music@pop,\; animal@cat)}} \\

The person who loves rock music is the tea drinker. &
\texttt{clue\_property\_at\_property(music@rock,\; drink@tea)} \\[4pt]

\bottomrule
\end{tabular}
\end{center}

\paragraph{Ambiguity}
The phrase \emph{“who has a cat”} of the 4th clue is semantically ambiguous in this domain:  
it could refer either to \emph{(i)} the person whose \textbf{pet} is a cat (\texttt{pet@cat}), or \emph{(ii)} the person who keeps the \textbf{animal} cat (\texttt{animal@cat}). The model selects \texttt{animal@cat}, but the ground-truth interpretation in the puzzle is \texttt{pet@cat}. 

Under the incorrect interpretation of the tool calls, this implies:
\[
\text{Due to clue 2 Eric has } \texttt{animal@cat} \quad\text{and}\quad \text{due to clue 4 Arnold has }\texttt{animal@cat},
\]
which violates the uniqueness constraint for animals.
Hence the set of constraints become unsatisfiable and \miniexact{} returns no solution.

This example demonstrates how category confusion can propagate into hard symbolic contradictions despite an otherwise valid puzzle.

\subsection{Example Failure due to Missing Initialization}\label{app:trace-DH} %
The following trace presents a representative failure caused by an error in the neural language module. While the puzzle is correctly harmonized and all clues are translated into symbolic constraints, the required \texttt{init\_zebra} tool call is omitted. As a result, the symbolic domain is never initialized and \miniexact{} cannot produce a solution, regardless of whether the tool calls were translated correctly.

\subsubsection{Puzzle Instance}

\begin{Verbatim}[breaklines=true,breakautoindent,fontsize=\footnotesize]
=== PUZZLE ===
There are 2 houses, numbered 1 to 2 from left to right, as seen
from across the street. Each house is occupied by a different person.
Each house has a unique attribute for each of the following characteristics:

- Each person has a unique name: `Arnold`, `Eric`
- The people keep unique animals: `cat`, `horse`
- People use unique phone models: `samsung galaxy s21`, `iphone 13`
- Each person has a unique type of pet: `cat`, `dog`
- Everyone has a unique favorite cigar: `pall mall`, `prince`

### Clues:

1. Arnold and the person who uses a Samsung Galaxy S21 are next to each other.
2. The person partial to Pall Mall and Eric are next to each other.
3. The person who owns a dog is directly left of Eric.
4. The cat lover and the person who uses a Samsung Galaxy S21 are next to each other.
   \end{Verbatim}

\subsubsection{Ground-Truth Solution}
\begin{center}
\begin{tabular}{c|ccccc}
\toprule
House & Name & Animal & Phone & Pet & Cigar \\
\midrule
1 & Arnold & cat   & iphone 13          & dog & pall mall \\
2 & Eric   & horse & samsung galaxy s21 & cat & prince \\
\bottomrule
\end{tabular}
\end{center}
\subsubsection{Expected Symbolic Domain Initialization}
Based on the harmonized puzzle, the language module should generate the following initialization call:
\begin{Verbatim}[breaklines=true,breakautoindent,fontsize=\footnotesize]
init_zebra(
    number_of_places = 2,
    number_of_clues = 4,
    property_groups = {
        name   = {name@arnold, name@eric},
        animal = {animal@cat, animal@horse},
        phone  = {phone@samsung-galaxy-s21, phone@iphone-13},
        pet    = {pet@cat, pet@dog},
        cigar  = {cigar@pall-mall, cigar@prince}
    }
)
\end{Verbatim}
However, the call itself is absent from the emitted tool call sequence. Because the symbolic domain is never initialized, \miniexact{} cannot construct the underlying constraint model. Consequently, no symbolic prediction is produced and the omission of the mandatory initialization call prevents the solver from executing and results in a complete failure to generate a solution.

\subsection{Example Failure due to Parsing Error}\label{app:trace-C} %

The following trace presents a representative failure caused by an error in the post-processing step that converts the symbolic solver output of \miniexact{} into the final structured solution. The symbolic constraints and solver result are correct and match the ground-truth solution, but the parsed solution misses an  attribute value.

\subsubsection{Puzzle Instance}

\begin{Verbatim}[breaklines=true,breakautoindent,fontsize=\footnotesize]
=== PUZZLE ===
There are 4 houses, numbered 1 to 4 from left to right, as seen 
from across the street. Each house is occupied by a different person. 
Each house has a unique attribute for each of the following characteristics:
 - Each person has a unique name: `Peter`, `Eric`, `Alice`, `Arnold`
 - People have unique hair colors: `brown`, `blonde`, `red`, `black`
 - People have unique heights: `short`, `average`, `tall`, `very short`
 - Each person has a unique favorite color: `green`, `yellow`, `white`, `red`

### Clues:
1. Eric is the person who has an average height.
2. The person whose favorite color is red is the person who is tall.
3. The person whose favorite color is green is somewhere to the left of Peter.
4. Arnold is in the fourth house.
5. The person who has an average height is not in the first house.
6. The person who is very short is in the fourth house.
7. The person who loves white is the person who has black hair.
8. The person who is tall is directly left of the person who has brown hair.
9. The person who has blonde hair is the person who has an average height.
\end{Verbatim}

\subsubsection{Ground-Truth Solution}
\begin{center}
\begin{tabular}{c|cccc}
\toprule
House & Name & Hair color & Height & Favorite color \\
\midrule
1 & Alice  & black  & short     & white \\
2 & Eric   & blonde & average   & green \\
3 & Peter  & red    & tall      & red \\
4 & Arnold & brown  & very short & yellow \\
\bottomrule
\end{tabular}
\end{center}

\subsubsection{Clues and Generated Tool Calls}
\begin{tabular}{p{0.2\linewidth} p{0.8\linewidth}}
\toprule
\textbf{Clue} & \textbf{Generated tool call} \\
\midrule
Clue 1 & \texttt{clue\_property\_at\_property(name@eric, height@average)} \\
Clue 2 & \texttt{clue\_property\_at\_property(favorite-color@red, height@tall)} \\
Clue 3 & \texttt{clue\_property\_left\_of(favorite-color@green, name@peter)} \\
Clue 4 & \texttt{clue\_property\_at\_house(name@arnold, 4)} \\
Clue 5 & \texttt{clue\_not\_property\_at\_house(height@average, 1)} \\
Clue 6 & \texttt{clue\_property\_at\_house(height@very-short, 4)} \\
Clue 7 & \texttt{clue\_property\_at\_property(favorite-color@white, hair-color@black)} \\
Clue 8 & \texttt{clue\_property\_directly\_left\_of\_property(height@tall, hair-color@brown)} \\
Clue 9 & \texttt{clue\_property\_at\_property(hair-color@blonde, height@average)} \\
\bottomrule
\end{tabular}

\subsubsection{\miniexact{} Output}

The symbolic solver returns a complete assignment consistent with the ground truth:

\begin{Verbatim}[breaklines=true,breakautoindent,fontsize=\footnotesize]
House 1:
  name = name@alice
  hair-color = hair-color@black
  height = height@short
  favorite-color = favorite-color@white

House 2:
  name = name@eric
  hair-color = hair-color@blonde
  height = height@average
  favorite-color = favorite-color@green

House 3:
  name = name@peter
  hair-color = hair-color@red
  height = height@tall
  favorite-color = favorite-color@red

House 4:
  name = name@arnold
  hair-color = hair-color@brown
  height = height@very-short
  favorite-color = favorite-color@yellow
\end{Verbatim}

This semantically corresponds exactly to the ground-truth solution.

\subsubsection{Parsed Prediction}
Parsing the \miniexact{} output into the final structured representation results in:

\begin{center}
\begin{tabular}{c|cccc}
\toprule
House & Name & Hair color & Height & Favorite color \\
\midrule
1 & Alice  & black  & short     & white \\
2 & Eric   & blonde & average   & green \\
3 & Peter  & red    & tall      & \bf{/} \\
4 & Arnold & brown  & very short & yellow \\
\bottomrule
\end{tabular}
\end{center}

The \texttt{favorite-color} value for house~3 (\texttt{red}) is lost during parsing, even though it is correctly present in the symbolic solver output. This indicates that the error is introduced entirely in the post-processing step, while symbolic reasoning remains correct.

We hypothesize that this failure occurs during the deharmonization step, in particular when multiple attributes share identical parts of their naming across different characteristics. In the illustrated example, the value \emph{red} appears as both a hair color and a favorite color, which may introduce ambiguity during matching. The current post-processing relies on approximate string matching between symbolic outputs and schema fields. Such overlaps can therefore lead to incorrect or missing assignments and resolving this suspected issue would require the implementation of a more robust disambiguation strategy.

\end{document}